%% file: acl_latex.tex
\documentclass[11pt]{article}

\usepackage[final]{acl}

\usepackage{times}
\usepackage{latexsym}

\usepackage{mathptmx}

\usepackage[T1]{fontenc}

\usepackage[utf8]{inputenc}

\usepackage{microtype}

\usepackage{inconsolata}

\usepackage{graphicx}

\usepackage{multirow}
\usepackage{amsmath}

\usepackage{booktabs}
\usepackage{siunitx}

\usepackage{color,soul} 
\usepackage{pifont}
\usepackage{inconsolata} 

\newcommand\blfootnote[1]{%
  \begingroup
  \renewcommand\thefootnote{}\footnote{#1}%
  \addtocounter{footnote}{-1}%
  \endgroup
}

\newcommand{\tscscore}[2]{\ensuremath{#1\,{\scriptstyle\pm\,#2}}}
\newcommand{\tscbest}[2]{\ensuremath{\mathbf{#1}\,{\scriptstyle\pm\,#2}}}

\providecommand{\tscscore}[2]{\ensuremath{#1\,{\scriptstyle\pm\,#2}}}
\providecommand{\tscbest}[2]{\ensuremath{\mathbf{#1}\,{\scriptstyle\pm\,#2}}}

\title{Translated Summary Conditioning for Multilingual \\ Large Language Model Pretraining}

\author{
Jiayi Wang$^{\alpha,\beta}$ \quad
Adam Vawda$^{\alpha,\beta}$ \quad
Xuyang Cao$^{\beta, \gamma}$ \quad
Qianying Liu$^{\beta}$ \quad
Chuang Ma$^{\beta, \delta}$ \\
\textbf{Chaoran Liu}$^{\beta}$ \quad
\textbf{Fei Chen}$^{\beta}$ \quad
\textbf{Shuyuan Zheng}$^{\gamma}$ \quad
\textbf{Hirokazu Kiyomaru}$^{\beta}$ \\
\textbf{Yao Lu}$^{\alpha}$ \quad
\textbf{Sadao Kurohashi}$^{\beta}$ \quad
\textbf{Pontus Stenetorp}$^{\alpha, \beta}$ \\
$^{\alpha}$University College London \quad 
$^{\beta}$LLMC,  National Institute of Informatics \quad \\
$^{\gamma}$The University of Osaka \quad
$^{\delta}$Kyoto University \\
\texttt{\{jiaywang,p.stenetorp\}@cs.ucl.ac.uk}
}

\begin{document}
\maketitle
\blfootnote{Jiayi Wang and Adam Vawda conducted this work while visiting LLMC, National Institute of Informatics.}

\begin{abstract}
Machine-translated data is widely used to compensate for the scarcity of the pretraining data of non-English large language models, yet it remains unclear what text should be translated and how it should be combined with non-translated data. We propose Translated Summary Conditioning (TSC), a simple pretraining data construction method that generates a one-sentence summary for each English document using an LLM, translates the summary into a target language, and prepends it to its English source document. This turns an English corpus into bilingual training examples, in which each non-English summary describes the English document that follows, while increasing the corpus size by only 4.2--22.4\%. We pretrain 1.3B-parameter models from scratch on five typologically diverse languages: French, Japanese, Hindi, Bengali, and Swahili. Each is compared with a baseline trained on the same English and native non-English documents without prepended summaries. TSC improves average non-English performance for all five languages, with average gains of up to 8.52 points for Japanese on our evaluation tasks, while English performance remains largely unchanged.\footnote{To support reproducibility, we will release the TSC corpora, trained models, and code under Open Source Initiative-approved licenses upon acceptance.}
\end{abstract}

\section{Introduction}
\label{sec:intro}
Although multilingual large language models~(LLMs) have made steady advancements in natural language processing~\citep{dubey2024llama, qwen2025qwen25technicalreport, team2024gemma}, their development remains limited by the scarcity of high-quality training data for the vast majority of languages~\citep{joshi2020state,kreutzer2022quality}.

A common response is to use synthetic data, particularly machine-translated text, to strengthen their multilingual capabilities~\citep{dubey2024llama,bornea2021multilingual,urbizu-etal-2023-enough,boughorbel2024improving,doshi2024translationese,wang2025multilingual}. \citet{wang2025multilingual} pretrained on approximately 1.6 trillion tokens of translated data from~\textit{FineWeb-Edu}~\citep{lozhkov2024fineweb-edu} and showed that the resulting model matched or outperformed multilingual LLMs of similar size, including~\textit{Qwen2.5}~\citep{qwen2025qwen25technicalreport} and~\textit{Gemma3}~\citep{team2025gemma}, on non-English understanding and reasoning tasks. At smaller scales, a 3-billion-token Spanish corpus translated into Basque~\citep{urbizu-etal-2023-enough}, and a corpus of 2.2 million English stories translated into Arabic~\citep{boughorbel2024improving} have been shown to benefit downstream tasks, such as natural language understanding and story generation. 

Despite these successes, existing studies differ widely in what they translate and how they combine it with natural data, making it difficult to draw general conclusions about how translated data should be used for pretraining.

Our work is motivated by two open research questions: (i)~\textit{Content:} which source data is most effective to translate? (ii)~\textit{Integration:} how should translated data be combined with non-translated web data during pretraining? While a comprehensive answer to both is beyond the scope of a single study, we take a step toward addressing them by drawing on two lines of prior work.
First, metadata conditioning shows that prepending short cues such as URLs to documents accelerates pretraining~\citep{gao2025metadata}. We hypothesise that a richer, more document-specific cue, such as a summary, might provide a more informative conditioning signal when prepended to the corresponding documents.
Second, recent studies show that mixed-language documents in pretraining data provide bilingual signals that contribute to the cross-lingual abilities of LLMs~\citep{briakou2023searching,blevins2022language,shao-etal-2026-role}. We therefore hypothesise that writing these summaries in non-English languages and prepending them to the corresponding English documents could strengthen multilingual capabilities.

Building on these insights, we propose \textit{Translated Summary Conditioning} (TSC). We use \textit{Gemma3} (12B) to summarise documents from the 100B-token subset of \textit{FineWeb-Edu}~\citep{lozhkov2024fineweb-edu}. 
We then translate these summaries into non-English languages with the neural machine translation model \textit{NLLB-200-1.3B}~\citep{costa2022no}, and prepend each translated summary to its corresponding English document. The resulting bilingual examples replace the English-only documents, while non-English web documents are retained. Unlike parallel or code-switched text, these examples are aligned at the document level, allowing us to examine whether such alignment benefits non-English capabilities.


We conduct experiments on five typologically diverse languages: French, Japanese, Hindi, Bengali, and Swahili, which span three language families, four writing systems, and resource levels from high to low~\citep{joshi2020state}. For each language, we pretrain 1.3B-parameter models from scratch using TSC and baseline corpora under the same training budget. Our results show that TSC improves average non-English performance for all five languages on commonsense reasoning and understanding tasks, with the largest gain for Japanese (+8.52 point average over our evaluation benchmarks), and outperforms the baseline on 14 of 18 non-English understanding tasks, while English performance remains largely unchanged. Ablations show that these gains do not come solely from additional non-English tokens: pairing translated summaries with their English source documents is crucial for Hindi and Bengali, improves Japanese commonsense reasoning, and improves French$\rightarrow$English translation. We also apply cross-lingual concept patching~\citep{dumas-etal-2025-separating,korner-etal-2026-meanings}, a mechanistic interpretability method measuring shared internal concept representations across multiple languages, to Japanese, Hindi, and Bengali. This analysis demonstrates that TSC exhibits stronger cross-lingual concept alignment than the ablation that adds the translated summaries as separate documents; this stronger alignment may stem from TSC seeing the same content in both languages within a single document during training.


Our contributions are as follows: \textbf{(1)}~We propose TSC, a simple method that prepends translated document summaries to their English sources, creating bilingual training examples at low translation cost; \textbf{(2)}~We show, through controlled experiments on five typologically diverse languages, that TSC improves non-English performance without degrading English; and \textbf{(3)}~We show through ablations that, for most languages, these gains depend on pairing non-English summaries with their English source documents rather than on the additional non-English tokens alone.

\section{Related Work}
\paragraph{Translated data for multilingual LLMs}
Machine translation is widely used to synthesise training data for languages with scarce native text~\citep{dubey2024llama,bornea2021multilingual,doshi2024translationese}. Translated corpora can substitute for and complement native data~\citep{urbizu-etal-2023-enough}, and remain effective at the scale of over a trillion tokens~\citep{wang2025multilingual}. However, translation can introduce errors and cultural biases that require additional high-quality native data to correct~\citep{boughorbel2024improving}. Unlike prior work, which translates entire documents as standalone training examples, we translate concise summaries and attach them to their English source documents.

\paragraph{Metadata conditioning in pretraining}
\citet{gao2025metadata} show that prepending metadata such as URLs to documents accelerates pretraining, allowing a 1.6B model to match standard pretraining with 33\% less data, and that model-generated topics can also serve as effective metadata. Such conditioning turns unconditional language modelling into conditional modelling, in which the model learns to predict text given a cheap, high-level cue about the document's source or content. Our work extends this idea from short metadata to document summaries, and from monolingual to cross-lingual conditioning.

\paragraph{Bilingual signals in pretraining}
Bilingual content present in web corpora has been linked to the cross-lingual abilities of LLMs~\citep{blevins2022language,briakou2023searching,cao2026shapleylaw}, and explicitly pairing aligned text across languages has long been used to encourage cross-lingual representations~\citep{conneau2019cross,qorib2025just}. However, recent controlled studies show that the benefit depends on the type of bilingual data: \citet{shao-etal-2026-role} find that parallel documents are critical for translation, whereas code-switched documents contribute little; cross-lingual understanding and reasoning remain largely unaffected by either. Our translated summaries occupy a middle ground between these categories: They are semantically aligned with the English document at the document level, but are neither sentence-level parallel text nor code-switched text.

\section{Translated Summary Conditioning Pretraining}
\label{sec:tsc}
\begin{figure*}
    \centering
    \includegraphics[width=0.8\textwidth]{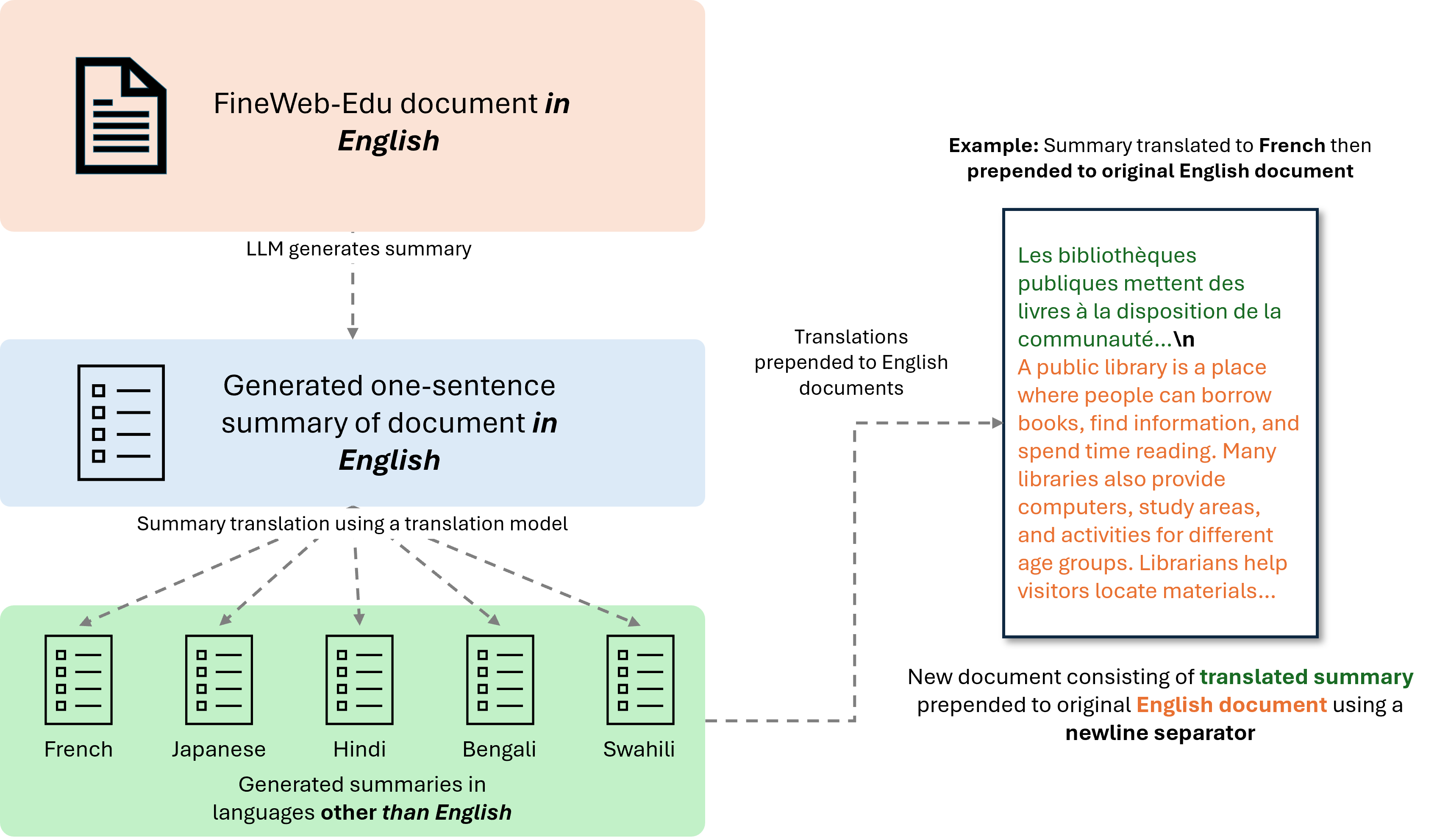}
    \caption{TSC generates a multilingual pretraining corpus by summarising high-quality English documents, translating these summaries, then prepending these translated summaries to the original English text.}
    \label{fig:method}
\end{figure*}
This section describes our pipeline for constructing a pretraining corpus with TSC and using it to pretrain multilingual LLMs. 

\subsection{Pretraining Data Curation}
\label{sec:data_curation}
We construct TSC training examples from four components: source data, a summarisation model, a summary translation model, and a strategy for composing bilingual examples.

\paragraph{Summary Generation}
We use a randomly sampled 100B-token subset of \textit{FineWeb-Edu}~\citep{lozhkov2024fineweb-edu}, a subset of FineWeb~\citep{penedo2024fineweb} filtered for educational value using automated quality annotations, whose effectiveness for pretraining has been previously demonstrated~\citep{Karpathy2024,martins2024eurollm,wang2025multilingual}.
We generate a one-sentence summary for each English document using \textit{Gemma3}~(12B)~\citep{team2025gemma} as shown in Figure~\ref{fig:summary-prompt} in Appendix~\ref{app:prompt}.
We run inference with vLLM~\citep{kwon2023efficient}, using nucleus sampling with a temperature of 0.6, top-$p$ of 0.95, and a maximum generation length of 512 tokens. Documents exceeding the model's context window, after reserving space for the prompt and the generated summary, are truncated before summarisation. We utilised 5,069 GPU hours on NVIDIA H200 GPUs to generate the summaries. On average, summaries contain 47 tokens, compared with 1,038 tokens on average for the source documents.

\paragraph{Summary Translation}
We select five target languages that span three language families, four writing systems, and a range of resource levels: French (fr; Indo-European, Latin script), Japanese (ja; Japonic, Japanese script), Hindi (hi; Indo-European, Devanagari script), Bengali (bn; Indo-European, Bengali script), and Swahili (sw; Niger-Congo, Latin script). Following the taxonomy of~\citet{joshi2020state}, French and Japanese are high-resource languages (Class~5), Hindi is a medium- to high-resource language (Class~4), Bengali is a medium-resource language (Class~3), and Swahili is a low-resource language (Class~2).
Translating summaries for a 100B-token corpus into multiple languages requires a translation model that is inexpensive to run at scale. We therefore adopt \textit{NLLB-200-1.3B}~\citep{costa2022no}, following \citet{wang2025multilingual}. Compared to LLM-based translators~\citep{alves2024tower,martins2024eurollm}, which are usually several times larger and tend to concentrate on a limited set of high-resource languages, NLLB-200 covers 200 languages at a fraction of the inference cost. Moreover, because each summary is a single sentence, it fits naturally with NLLB's sentence-level design and can be translated directly, without the segmentation and reassembly required for document translation. Translation is run with a batch size of 64 and a beam size of 3, requiring 960 GPU hours per language on NVIDIA H200 GPUs.

\input{Experiments/tables/corpus-statistics}
\paragraph{Composing the TSC corpus}
For each English document, we prepend its translated summary in a target language, separated by a newline, as illustrated in Figure~\ref{fig:method}.
Unlike parallel or code-switched data, each resulting example pairs a non-English summary with a semantically aligned, but not translation-equivalent, English document. We then combine these bilingual examples with native non-English documents from \textit{FineWeb2}~\citep{penedo2025fineweb2pipelinescale} to form the TSC pretraining corpus. We choose \textit{FineWeb2} because it extends the \textit{FineWeb} processing pipeline, from which our English source data is derived, to over 1,000 languages with language-specific filtering and deduplication, and has been shown to yield stronger multilingual models than prior web corpora. In contrast, our baseline corpus contains the same native non-English documents, combined with the original English documents without prepended summaries. Table~\ref{tab:token-statistics} reports the total size of the baseline and TSC corpora for each language pair, and Table~\ref{tab:monolingual-token-counts} (Appendix~\ref{app:corpus-composition}) breaks down the English and native non-English data they share. Prepending translated summaries increases the corpus size by only 4.2\% to 22.4\% across language pairs, with the largest increase for Bengali, likely due to the higher token fertility of the Bengali script.


\subsection{Multilingual LLM Pretraining}
\label{sec:pretraining}
To isolate the effect of TSC, we train one model per language pair on the TSC corpus and one on the corresponding baseline corpus, keeping the architecture, tokeniser, and optimisation settings identical within each pair.

\paragraph{Architecture and optimisation}
All models are 1.3B-parameter decoder-only Transformers whose design follows the Llama family~\citep{touvron2023llama2} and open GPT-2 reproductions~\citep{Karpathy2024}, using rotary position embeddings, SwiGLU activations, and RMSNorm. Training uses sequences of 2,048 tokens and a global batch of 2,048 sequences, corresponding to roughly 4M tokens per step, with a constant learning rate of $2 \times 10^{-4}$. Appendix~\ref{sec:appendix-model-arch-params} lists the complete configuration.

\paragraph{Tokenisation}
The French, Japanese, and Swahili models use the Llama~2 tokeniser, consistent with \citet{wang2025multilingual}, while the Hindi and Bengali models use the Llama~3 tokeniser for its broader coverage of Indic scripts~\citep{dubey2024llama}.

\paragraph{Data sampling}
Training sequences are formed by randomly drawing documents from the pretraining corpus and packing them, separated by end-of-sequence tokens, up to the sequence length. Each TSC example, consisting of a translated summary followed by its English source document, is treated as a single document, so the summary and the document it describes always appear contiguously in the same sequence.

\paragraph{Implementation}
Models are trained from scratch with Megatron-LM~\citep{shoeybi2019megatron} and FlashAttention-2~\citep{dao2023flashattention} on NVIDIA H200 GPUs. Each model is trained for 140{,}000 steps (about 587B tokens), using approximately 3,200 H200 GPU hours.

\input{Experiments/Experiments}

\input{Experiments/Ablations}

\input{mi_part/mi_concept_readout}   

\section{Conclusion}
\label{sec:conclusion}

We introduced Translated Summary Conditioning~(TSC), a pretraining data construction method that prepends machine-translated, LLM-generated summaries to their English source documents, forming document-level aligned bilingual training examples. Across the five typologically diverse languages, TSC improves the average non-English performance over a controlled baseline trained on the same English and native non-English data, while English performance remains largely unchanged. 
Our ablations indicate that these gains do not stem solely from additional non-English tokens: pairing the translated summaries with their English sources is crucial for Hindi and Bengali, and further improves Japanese JCommonsenseQA and XWinograd as well as French$\rightarrow$English translation, whereas prepending English summaries alone has no meaningful effect.
Cross-lingual concept patching further shows that TSC aligns concepts between English and Japanese, Hindi, and Bengali more strongly than adding the same summaries as standalone documents, suggesting that seeing the same content in both languages within a single document strengthens cross-lingual alignment.



\section*{Limitations}
While TSC shows promising results for improving multilingual capabilities in pretraining, our study has several limitations that we hope future work will address.

\paragraph{Model scale} Due to computational constraints, all experiments use 1.3B-parameter models. It therefore remains an open question whether the benefits of TSC persist, grow, or diminish at larger scales.

\paragraph{Language coverage} Our study covers five languages, chosen to span different language families, writing systems, and resource levels. Although this selection is diverse, it is necessarily limited; extending the evaluation to languages from other regions and families (such as those of South America and Oceania) would help assess how well our findings generalize.


\paragraph{Ablation scope} Our English-summary ablation is conducted in a monolingual setting. A multilingual counterpart, in which English summaries are prepended to English documents while the native non-English data is retained, would further clarify the role of summary conditioning in the gains of TSC.

\section*{Acknowledgments}
Claude and ChatGPT were used to correct grammatical errors during the writing of the paper.

\bibliography{custom}

\appendix
\section{Summary Generation Prompt}
\label{app:prompt}
We generate a one-sentence summary for each English document with \textit{Gemma3} (12B), using the prompt in Figure~\ref{fig:summary-prompt}, where \texttt{\{doc\}} is replaced by the document text. Decoding uses nucleus sampling with a temperature of 0.6, top-$p$ of 0.95, and a maximum generation length of 512 tokens.

\begin{figure}[h]
\centering
\small
\begin{tabular}{@{}p{0.95\columnwidth}@{}}
\toprule
\begin{verbatim}
Summarize the document in one sentence.

Rules:
- Do NOT use bullet points
- Do NOT mention that this is a summary
- Use only one sentence

Document:
[/BEGIN_DOC]
{doc}
[/END_DOC]

Summary:
\end{verbatim}
\\
\bottomrule
\end{tabular}
\caption{Prompt used to generate document summaries with \textit{Gemma3} (12B).}
\label{fig:summary-prompt}
\end{figure}

\section{Composition of the Pretraining Corpora}
\label{app:corpus-composition}
Table~\ref{tab:monolingual-token-counts} reports the number of tokens of the English and native non-English data used in each language pair. The baseline and TSC corpora for a given language pair contain exactly the same native non-English documents from \textit{FineWeb2}~\citep{penedo2025fineweb2pipelinescale}; they differ only in whether translated summaries are prepended to the English documents. For the high-resource languages, French and Japanese, we sample a number of \textit{FineWeb2} tokens comparable to the English data. For Hindi, Bengali, and Swahili, where the available data is much smaller, we use all \textit{FineWeb2} data for the language. The table is divided into two groups according to the tokeniser used: following \citet{wang2025multilingual}, the French, Japanese, and Swahili experiments use the Llama~2 tokeniser, while the Hindi and Bengali experiments use the Llama~3 tokeniser for its better coverage of the Devanagari and Bengali scripts. The English data is the same 100B-token subset of \textit{FineWeb-edu} in all settings, so the two English counts differ only because of the tokeniser; token counts are therefore comparable within each group but not across groups.

\begin{table}[h]
    \centering
    \small
    \setlength{\tabcolsep}{12pt}
    \renewcommand{\arraystretch}{1.1}
    \begin{tabular}{@{} l S[table-format=3.1] @{}}
        \toprule
        Language & {Tokens (B)} \\
        \midrule
        \multicolumn{2}{@{}l}{\textit{Llama~2 tokeniser}} \\
        English (en)  & 111.4 \\
        French (fr)   &  86.7 \\
        Japanese (ja) &  86.4 \\
        Swahili (sw)  &   1.4 \\
        \midrule
        \multicolumn{2}{@{}l}{\textit{Llama~3 tokeniser}} \\
        English (en)  &  94.5 \\
        Hindi (hi)    &  25.8 \\
        Bengali (bn)  &  41.3 \\
        \bottomrule
    \end{tabular}
    \caption{Token counts (in billions) of the English and native non-English data shared by the baseline and TSC corpora, grouped by the tokeniser used. Counts are comparable within each group but not across groups.}
    \label{tab:monolingual-token-counts}
\end{table}

\section{Hyperparameters Settings for Model Pretraining}
\label{sec:appendix-model-arch-params}
\input{Experiments/tables/model-parameters}
Pretraining hyperparameter settings are shown in Table~\ref{tab:model-arch}.

\section{Evaluation Benchmarks for
Each Language}
\label{sec:appendix-benchmarks}

\input{Experiments/tables/benchmarks}
Specific evaluation benchmarks for each language are shown in Table~\ref{tab:benchmark_categories}.

\section{Per-Task Training Curves}
\label{app:training-curves}
\begin{figure*}[t]
    \centering
    \includegraphics[width=0.9\textwidth]{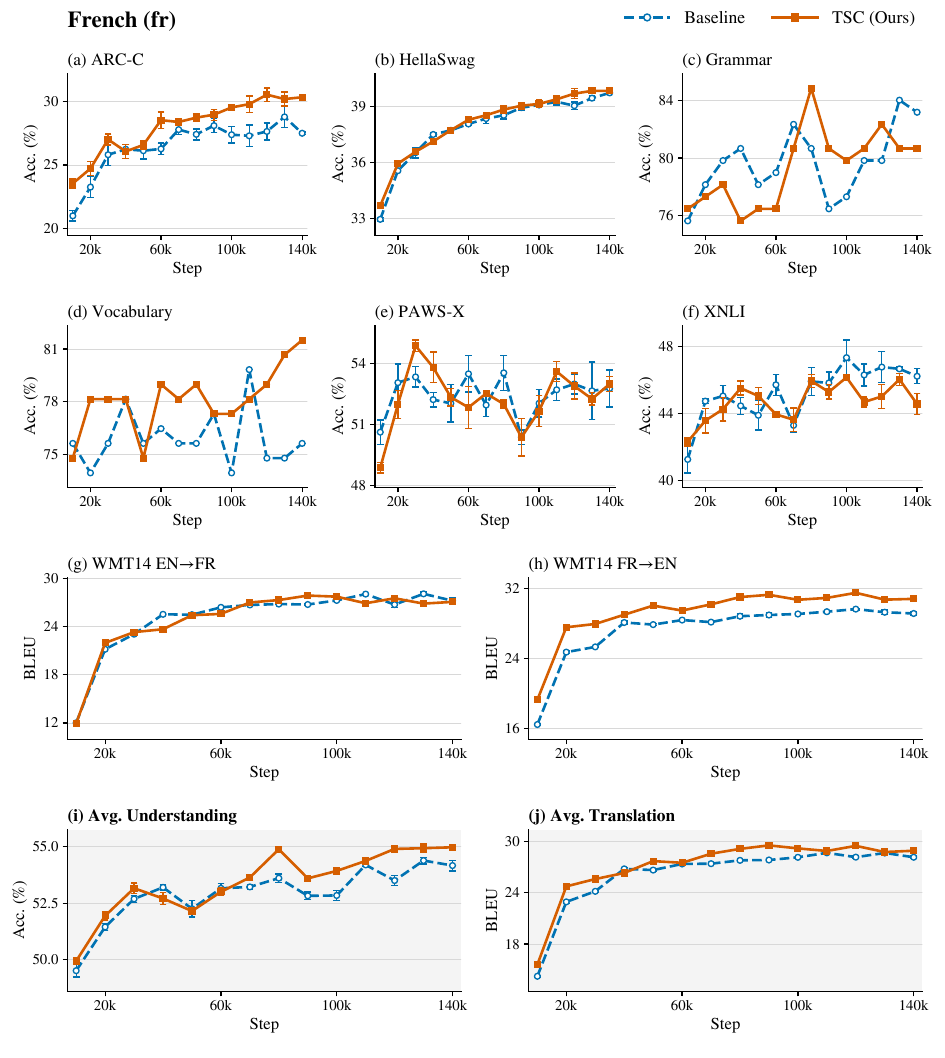}
    \caption{Training curves for French understanding and translation tasks.}
    \label{fig:curves-fr}
\end{figure*}

\begin{figure*}[t]
    \centering
    \includegraphics[width=0.9\textwidth]{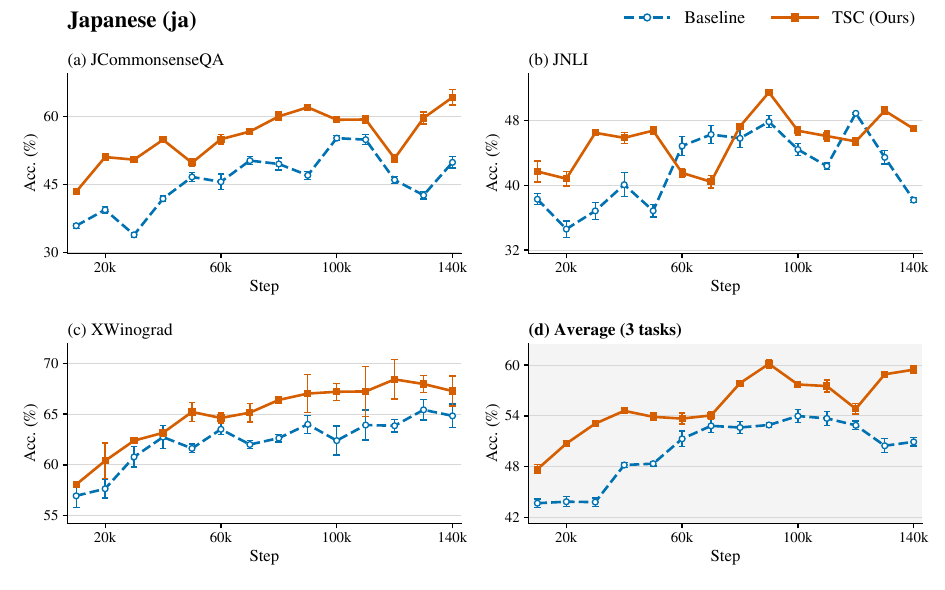}
    \caption{Training curves for Japanese tasks.}
    \label{fig:curves-ja}
\end{figure*}

\begin{figure*}[t]
    \centering
    \includegraphics[width=0.9\textwidth]{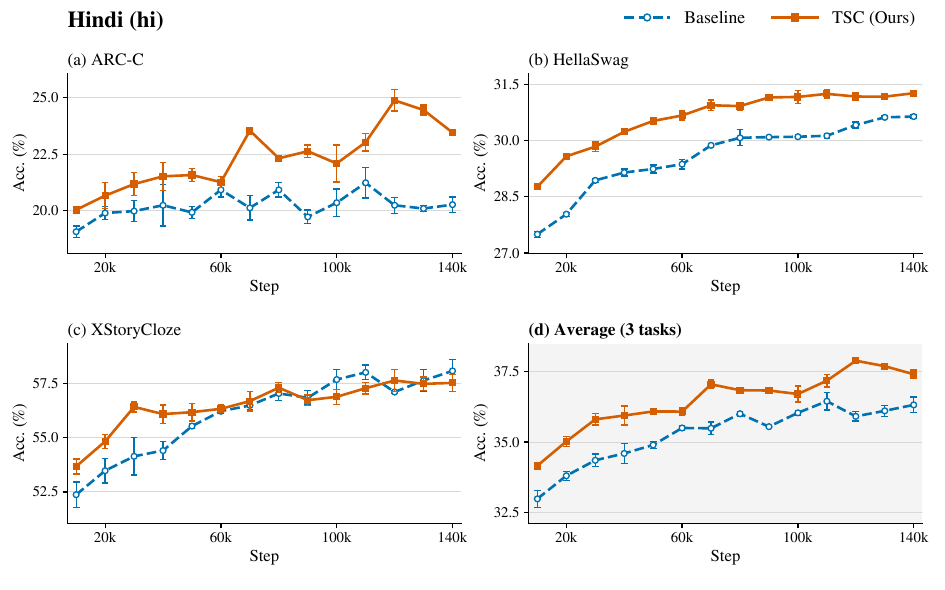}
    \caption{Training curves for Hindi tasks.}
    \label{fig:curves-hi}
\end{figure*}

\begin{figure*}[t]
    \centering
    \includegraphics[width=0.9\textwidth]{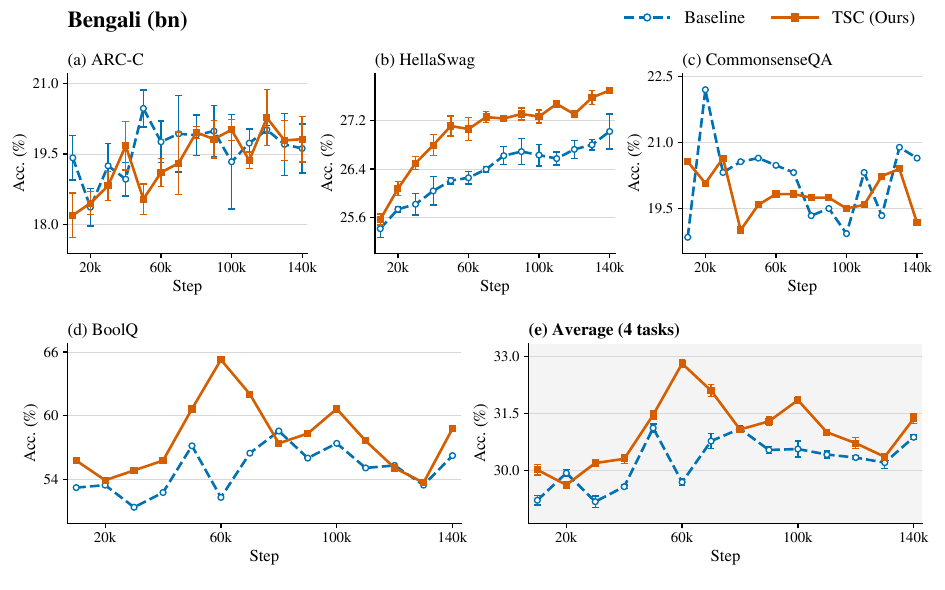}
    \caption{Training curves for Bengali tasks.}
    \label{fig:curves-bn}
\end{figure*}

\begin{figure*}[t]
    \centering
    \includegraphics[width=0.9\textwidth]{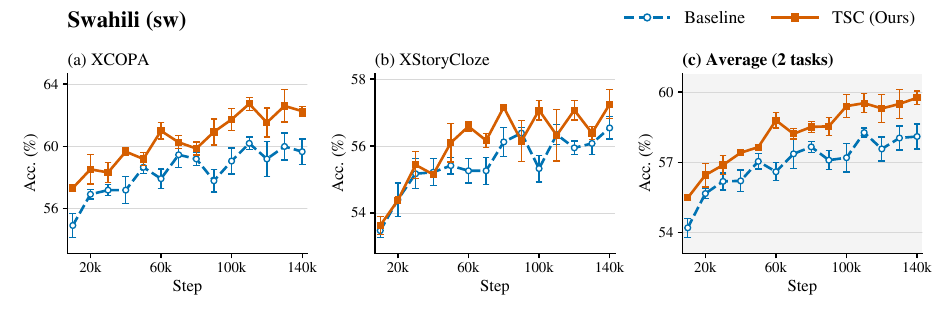}
    \caption{Training curves for Swahili tasks.}
    \label{fig:curves-sw}
\end{figure*}

Figures~\ref{fig:curves-fr}--\ref{fig:curves-sw} show the training curves of the TSC and baseline models for each benchmark, evaluated every 10{,}000 steps up to 140{,}000 steps. Languages are ordered by resource level, from high to low. Each point reports the mean over three evaluation seeds, and error bars indicate the standard deviation.

\section{English Evaluation Results}
\label{app:english-results}
Table~\ref{tab:tsc-140k-english} reports English results for each TSC model and its corresponding baseline after 140{,}000 training steps. All models are evaluated on the same seven English tasks (ARC-Easy, ARC-Challenge, HellaSwag, PIQA, PAWS-X, XNLI, and XWinograd) in a 5-shot setting, and we report accuracy averaged over three evaluation seeds.

\input{Experiments/tables/eval-results-EN}


\section{Per-Task Training Curves for the No-Prepend Ablation}
\label{app:abla-curves}
\begin{figure*}[t]
    \centering
    \includegraphics[width=0.9\textwidth]{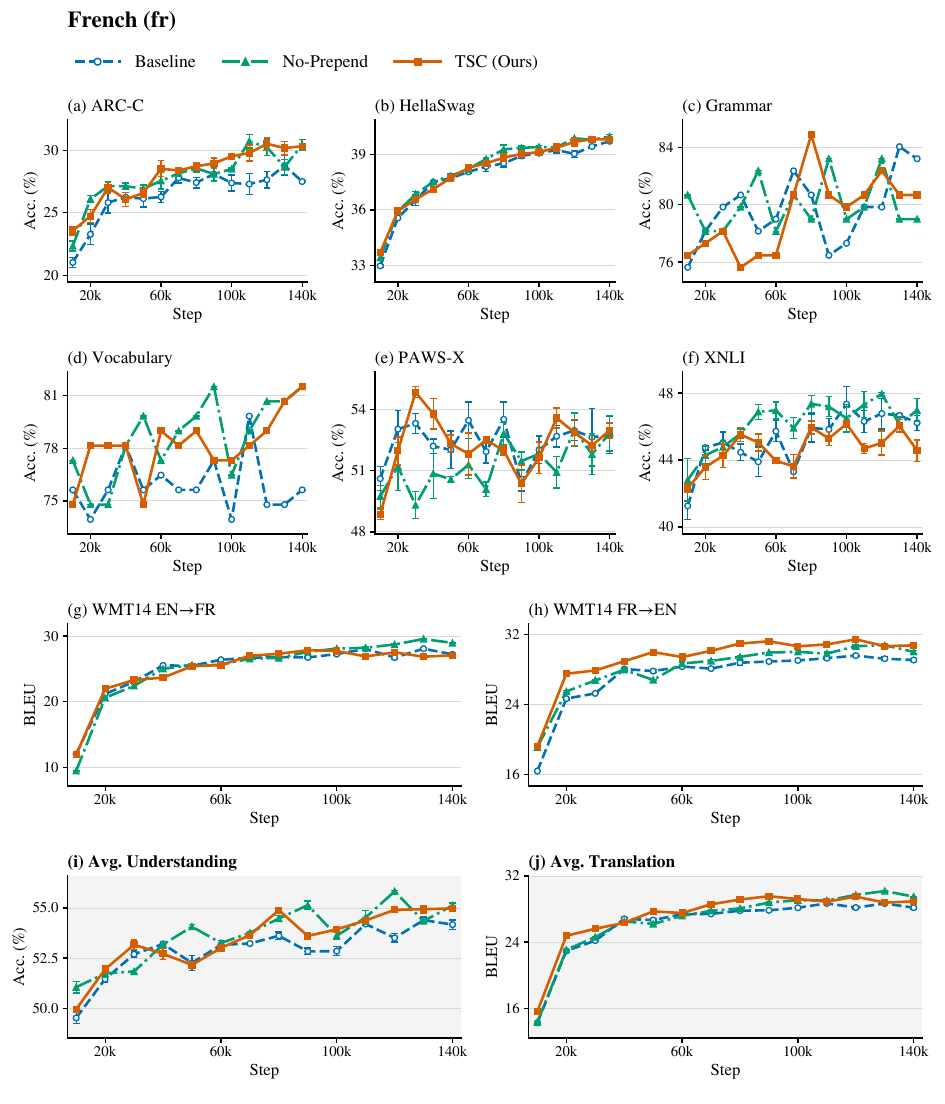}
    \caption{No-Prepend ablation: Training curves for French understanding and translation tasks.}
    \label{fig:abla-curves-fr}
\end{figure*}

\begin{figure*}[t]
    \centering
    \includegraphics[width=0.9\textwidth]{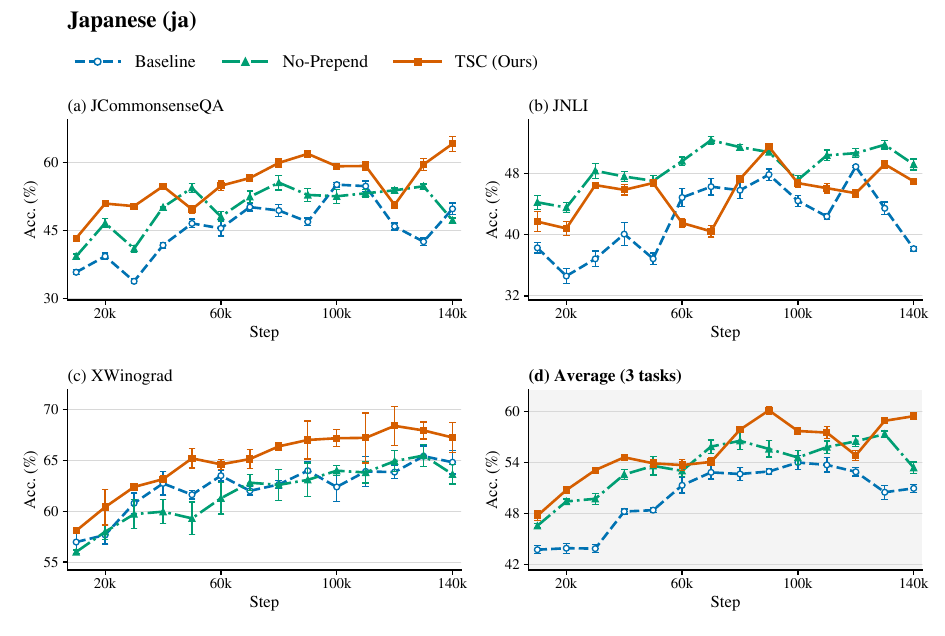}
    \caption{No-Prepend ablation: Training curves for Japanese tasks.}
    \label{fig:abla-curves-ja}
\end{figure*}

\begin{figure*}[t]
    \centering
    \includegraphics[width=0.9\textwidth]{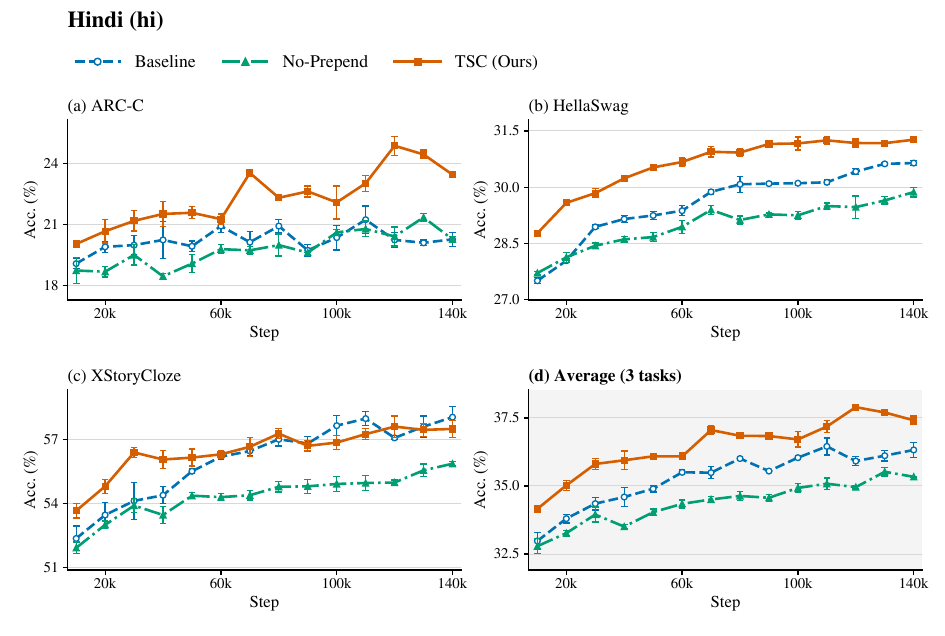}
    \caption{No-Prepend ablation: Training curves for Hindi tasks.}
    \label{fig:abla-curves-hi}
\end{figure*}

\begin{figure*}[t]
    \centering
    \includegraphics[width=0.9\textwidth]{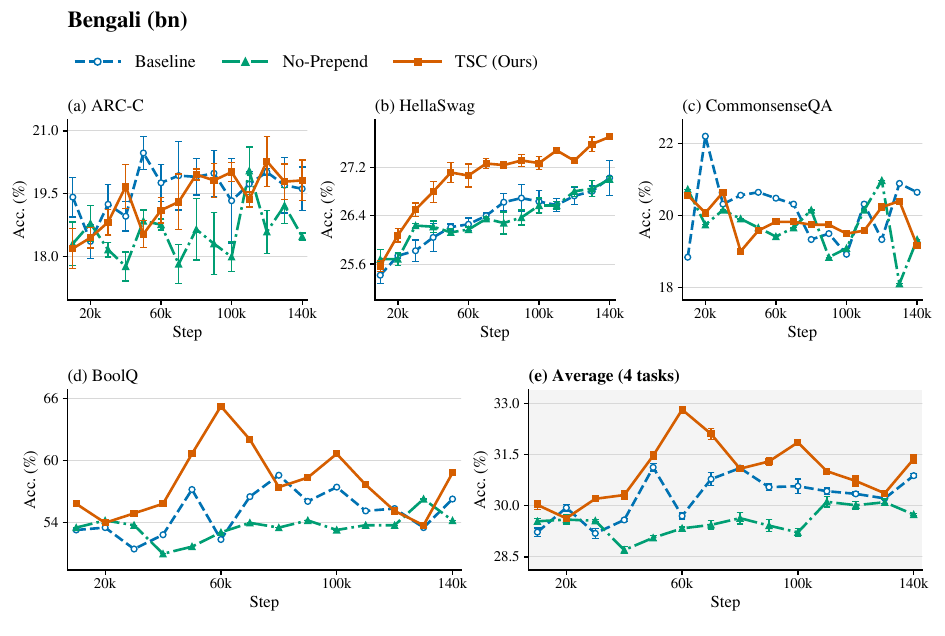}
    \caption{No-Prepend ablation: Training curves for Bengali tasks.}
    \label{fig:abla-curves-bn}
\end{figure*}

\begin{figure*}[t]
    \centering
    \includegraphics[width=0.9\textwidth]{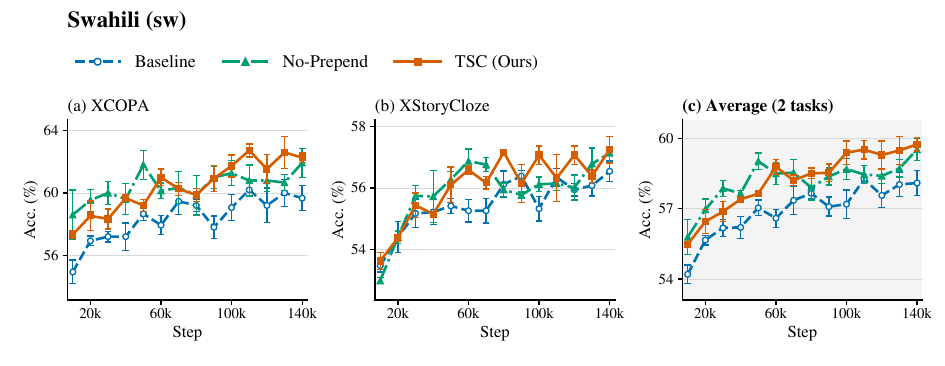}
    \caption{No-Prepend ablation: Training curves for Swahili tasks.}
    \label{fig:abla-curves-sw}
\end{figure*}

Figures~\ref{fig:abla-curves-fr}--\ref{fig:abla-curves-sw} compare the training curves of the baseline, No-Prepend, and TSC models, evaluated every 10{,}000 steps up to 140{,}000 steps. 
%
The figures show per-task curves for each language, ordered by resource level from high to low. Each point reports the mean over three evaluation seeds, and error bars indicate the standard deviation.


\section{English-Summary Ablation Results}
\label{app:abla-en}
To test whether summary conditioning alone, without translation, affects performance, we train a model on English documents with their English summaries prepended (\S\ref{sec:abla-en-summary}), and compare it with the baseline trained on the original English documents. Table~\ref{tab:tsc-140k-ablation-en} reports results on the English evaluation benchmarks after 140{,}000 training steps, averaged over three evaluation seeds.

\input{Experiments/tables/ablation-en}

\input{mi_part/mi_appendix} 

\end{document}

%% file: Experiments/tables/corpus-statistics.tex

\begin{table}[t]
    \centering
    \small
    \setlength{\tabcolsep}{5pt}
    \renewcommand{\arraystretch}{1.1}
    \begin{tabular}{
        @{} l
        S[table-format=3.1]
        S[table-format=3.1]
        S[table-format=2.1]
        @{}
    }
        \toprule
        Pair
        & {Baseline (B)}
        & {TSC (B)}
        & {Incr. (\%)} \\
        \midrule
        English--French & 198.1 & 206.4 &  4.2 \\
        English--Japanese & 197.8 & 210.6 &  6.5 \\
        English--Hindi & 120.3 & 133.3 & 10.8 \\
        English--Bengali & 135.8 & 166.2 & 22.4 \\
        English--Swahili & 112.8 & 123.6 &  9.6 \\
        \bottomrule
    \end{tabular}
    \caption{
        Token counts for the Baseline and TSC corpora per language pair and corresponding increases. 
    }
    \label{tab:token-statistics}
\end{table}

%% file: Experiments/Experiments.tex
\section{Experiments}
This section presents our evaluation of the performance of TSC pretraining across various multilingual benchmarks.

\subsection{Evaluation Benchmarks}

We evaluate the models trained on the TSC and baseline corpora on benchmarks in each of the five target languages and in English. The benchmarks span six categories: commonsense reasoning, sentence-pair understanding, reading comprehension, knowledge-based question answering, vocabulary and grammar, and translation; the full list is given in Table~\ref{tab:benchmark_categories} (Appendix~\ref{sec:appendix-benchmarks}). All evaluations use a 5-shot setting, and we report accuracy averaged over three random seeds.\footnote{For the translation tasks, we report BLEU scores.}

\subsection{Baselines}
\label{sec:baselines}
For each language pair, we compare the TSC model against a baseline model trained on the baseline corpus described in \S\ref{sec:data_curation}, which contains the same native non-English documents from \textit{FineWeb2} and the same English documents from \textit{FineWeb-Edu}, but without the prepended translated summaries. The two models share the same architecture, tokeniser, hyperparameters, and training budget of 140{,}000 steps (\S\ref{sec:pretraining}), so they differ only in whether English documents are prepended by translated summaries. 
\input{Experiments/tables/eval-results-140k}





\subsection{Main Results}
\label{sec:results}
\begin{figure*}[t]
    \centering
    \includegraphics[width=0.8\textwidth]{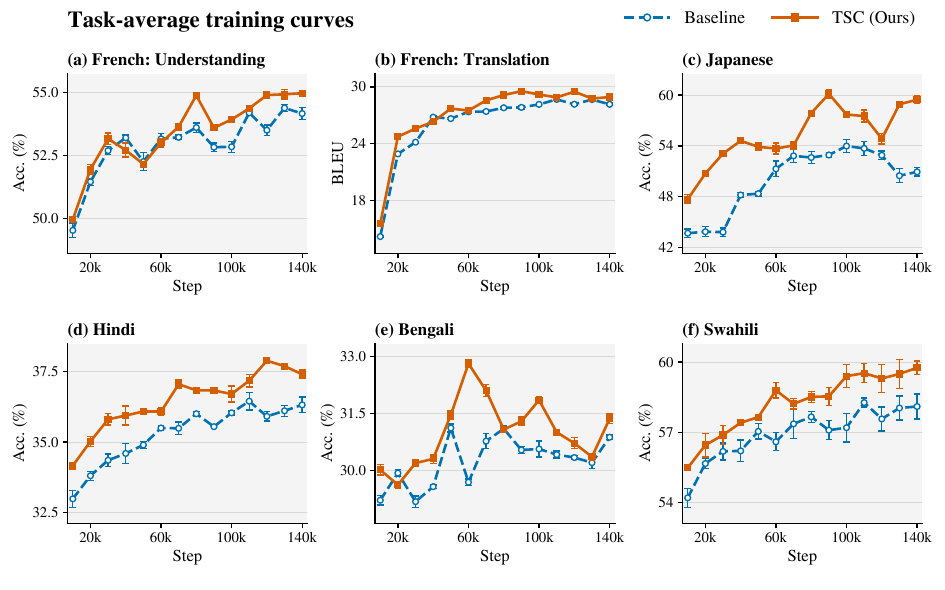}
    \caption{Task-average training curves through 140{,}000 steps.}
    \label{fig:eval_average}
\end{figure*}

Table~\ref{tab:tsc-140k} reports the results after 140{,}000 training steps, and Figure~\ref{fig:eval_average} shows the task-average performance throughout training; per-task training curves for each language are provided in Appendix~\ref{app:training-curves}. TSC improves the average performance over the baseline for all five target languages, with a mean gain of 2.51 points across languages, and outperforms the baseline on 14 of the 18 understanding tasks (15 of 20 when translation is included).

\paragraph{High-resource languages}
As shown in~Table~\ref{tab:tsc-140k}, for French, TSC yields a modest average gain of 0.81 points on understanding tasks. It improves ARC-C (+2.82) and Vocabulary (+5.88), but slightly decreases Grammar (-2.52) and XNLI (-1.66); because the FrenchBench Grammar and Vocabulary sets are small, these differences correspond to only a few questions. 
For translation, TSC improves French$\rightarrow$English by 1.67 BLEU, whereas English$\rightarrow$French remains essentially unchanged (-0.14). We attribute this improvement to the bilingual structure of TSC examples: Each example pairs a non-English summary with the English document it describes, forming a naturally occurring comparable pretraining-scale corpus. Comparable corpora have long been shown to benefit machine translation~\citep{munteanu-marcu-2005-improving}, and bilingual signals in pretraining data, particularly cross-lingually aligned content, have been linked to the translation capabilities of language models~\citep{kale-etal-2021-nmt5,briakou2023searching,shao-etal-2026-role}. The asymmetry between directions is further consistent with the ordering of TSC examples, in which the model learns to generate English text conditioned on preceding non-English content.
Japanese shows by far the largest improvement, with an average gain of 8.52 points driven by JCommonsenseQA (+14.33) and JNLI (+8.79). Although both French and Japanese are high-resource languages, Japanese is typologically more distant from English and uses a non-Latin script, suggesting that the semantic alignment provided by TSC is most valuable when there is little surface overlap with English.

\paragraph{Medium- and low-resource languages}
TSC also improves all three lower-resource languages on average across multiple benchmarks. For Hindi, a medium- to high-resource language~(Class~4), TSC increases the average by 1.09 points, driven mainly by ARC-C~(+3.20) and HellaSwag~(+0.62), with a slight drop on XStoryCloze~($-0.55$). For Bengali, the average gain is smaller~(+0.49): TSC improves BoolQ~(+2.55) and HellaSwag~(+0.68) but decreases CommonsenseQA~(-1.47). For Swahili, the lowest-resource language in our study, TSC improves both XCOPA~(+2.60) and XStoryCloze~(+0.71), for an average gain of 1.65 points. These results indicate that the benefits of TSC extend beyond high-resource languages.

\paragraph{Training dynamics}
The training curves in Figure~\ref{fig:eval_average} show that the advantage of TSC is not limited to the final checkpoint. For Japanese and Hindi, TSC outperforms the baseline at every checkpoint, starting from the first evaluation at 10{,}000 steps, and Swahili shows a smaller but similarly persistent benefit. Bengali is noisier, but TSC remains ahead at most checkpoints. For French, the understanding curves largely overlap, whereas TSC maintains a small but steady advantage in translation early in training. The early gap suggests that TSC accelerates the acquisition of non-English capabilities, in line with findings that conditioning documents on prepended information speeds up pretraining~\citep{gao2025metadata}.

\paragraph{Per-task trends}
The per-task curves (Figure~\ref{fig:curves-fr} to Figure~\ref{fig:curves-sw} in Appendix~\ref{app:training-curves}) show that the gains are concentrated on a few tasks where TSC leads throughout training: JCommonsenseQA for Japanese (often by 5--15 points), ARC-C and HellaSwag for Hindi, XCOPA for Swahili, HellaSwag for Bengali, and ARC-C and French$\rightarrow$English translation for French. For the remaining tasks, the two models largely overlap or fluctuate between checkpoints, particularly on the small FrenchBench sets and for Japanese JNLI, where the baseline drops at the final checkpoint.

Overall, TSC consistently improves non-English performance across languages with different scripts, families, and resource levels, although the magnitude of the gains varies considerably across languages and tasks, with the largest improvements for commonsense and knowledge-intensive tasks such as JCommonsenseQA and XCOPA for Swahili.

\paragraph{Performance in English}
Prepending translated summaries changes how the model sees every English document, thus we also check whether TSC affects English capabilities. Table~\ref{tab:tsc-140k-english} (Appendix~\ref{app:english-results}) reports English results for each TSC model and its baseline. The differences in average English accuracy are small, ranging from $-0.95$ (Bengali) to $+0.41$ (Hindi), with a mean change of $-0.27$ points across languages. TSC slightly improves the English average for Japanese and Hindi and slightly reduces it for French, Bengali, and Swahili. At the task level, TSC outperforms the baseline for 17 of the 35 task--language combinations, and the direction of the change for a given task is not consistent across language pairs. The largest drops occur on PAWS-X for the French model ($-4.13$) and ARC-C for the Bengali model ($-3.30$), while other models improve on the same tasks (e.g., Hindi on PAWS-X, $+1.45$; French on ARC-C, $+0.85$). These results suggest that TSC does not cause a systematic degradation of English performance: The multilingual gains reported above come with English capabilities that remain largely on par with the baseline.

%% file: Experiments/tables/eval-results-140k.tex
\begin{table}[t]
\centering
\small
\setlength{\tabcolsep}{3pt}
\renewcommand{\arraystretch}{0.9}
\begin{tabular*}{\columnwidth}{@{\extracolsep{\fill}}lrrr@{}}
\toprule
Task & Baseline & TSC (Ours) & $\Delta$ \\
\midrule
\multicolumn{4}{@{}l}{\textbf{French}} \\
\multicolumn{4}{@{}l}{\textit{Understanding (Acc.)}} \\
ARC-C & \tscscore{27.49}{0.10} & \tscbest{30.31}{0.13} & \tscscore{+2.82}{0.15} \\
HellaSwag & \tscscore{39.73}{0.05} & \tscbest{39.84}{0.07} & \tscscore{+0.11}{0.04} \\
Grammar & \tscbest{83.19}{0.00} & \tscscore{80.67}{0.00} & \tscscore{-2.52}{0.00} \\
Vocabulary & \tscscore{75.63}{0.00} & \tscbest{81.51}{0.00} & \tscscore{+5.88}{0.00} \\
PAWS-X & \tscscore{52.77}{0.91} & \tscbest{52.98}{0.37} & \tscscore{+0.22}{0.58} \\
XNLI & \tscbest{46.22}{0.45} & \tscscore{44.56}{0.64} & \tscscore{-1.66}{1.07} \\
\addlinespace[2pt]
\textbf{Average} & \tscscore{54.17}{0.24} & \tscbest{54.98}{0.08} & \tscscore{+0.81}{0.29} \\
\addlinespace[3pt]
\multicolumn{4}{@{}l}{\textit{Translation: WMT14 (BLEU)}} \\
EN$\to$FR & \tscbest{27.21}{0.33} & \tscscore{27.07}{0.10} & \tscscore{-0.14}{0.34} \\
FR$\to$EN & \tscscore{29.12}{0.16} & \tscbest{30.79}{0.07} & \tscscore{+1.67}{0.14} \\
\addlinespace[2pt]
\textbf{Average} & \tscscore{28.17}{0.15} & \tscbest{28.93}{0.08} & \tscscore{+0.76}{0.11} \\
\midrule
\multicolumn{4}{@{}l}{\textbf{Japanese} \textit{(Acc.)}} \\
JCSQA & \tscscore{49.84}{1.25} & \tscbest{64.16}{1.67} & \tscscore{+14.33}{2.54} \\
JNLI & \tscscore{38.17}{0.32} & \tscbest{46.96}{0.29} & \tscscore{+8.79}{0.25} \\
XWinograd & \tscscore{64.82}{1.16} & \tscbest{67.26}{1.48} & \tscscore{+2.43}{2.31} \\
\addlinespace[2pt]
\textbf{Average} & \tscscore{50.94}{0.48} & \tscbest{59.46}{0.42} & \tscscore{+8.52}{0.82} \\
\midrule
\multicolumn{4}{@{}l}{\textbf{Hindi} \textit{(Acc.)}} \\
ARC-C & \tscscore{20.26}{0.35} & \tscbest{23.46}{0.09} & \tscscore{+3.20}{0.40} \\
HellaSwag & \tscscore{30.64}{0.06} & \tscbest{31.27}{0.01} & \tscscore{+0.62}{0.05} \\
XStoryCloze & \tscbest{58.06}{0.51} & \tscscore{57.51}{0.40} & \tscscore{-0.55}{0.73} \\
\addlinespace[2pt]
\textbf{Average} & \tscscore{36.32}{0.28} & \tscbest{37.41}{0.15} & \tscscore{+1.09}{0.33} \\
\midrule
\multicolumn{4}{@{}l}{\textbf{Bengali} \textit{(Acc.)}} \\
ARC-C & \tscscore{19.62}{0.52} & \tscbest{19.82}{0.49} & \tscscore{+0.20}{0.05} \\
HellaSwag & \tscscore{27.02}{0.29} & \tscbest{27.70}{0.01} & \tscscore{+0.68}{0.28} \\
CSQA & \tscbest{20.64}{0.00} & \tscscore{19.16}{0.00} & \tscscore{-1.47}{0.00} \\
BoolQ & \tscscore{56.25}{0.00} & \tscbest{58.80}{0.00} & \tscscore{+2.55}{0.00} \\
\addlinespace[2pt]
\textbf{Average} & \tscscore{30.88}{0.06} & \tscbest{31.37}{0.12} & \tscscore{+0.49}{0.06} \\
\midrule
\multicolumn{4}{@{}l}{\textbf{Swahili} \textit{(Acc.)}} \\
XCOPA & \tscscore{59.67}{0.81} & \tscbest{62.27}{0.31} & \tscscore{+2.60}{0.53} \\
XStoryCloze & \tscscore{56.54}{0.33} & \tscbest{57.25}{0.43} & \tscscore{+0.71}{0.14} \\
\addlinespace[2pt]
\textbf{Average} & \tscscore{58.10}{0.53} & \tscbest{59.76}{0.29} & \tscscore{+1.65}{0.27} \\
\bottomrule
\end{tabular*}
\caption{Results at 140{,}000 steps (mean $\pm$ sample SD over three evaluation seeds). 
$\Delta$: seed-matched TSC $-$ Baseline. Bold: higher displayed mean.}
\label{tab:tsc-140k}
\label{tab:tsc-140k-fr}
\label{tab:tsc-140k-ja}
\label{tab:tsc-140k-hi}
\label{tab:tsc-140k-bn}
\label{tab:tsc-140k-sw}
\end{table}

%% file: Experiments/Ablations.tex
\section{Ablations and Discussion}



\label{sec:ablation}
To understand where the gains of TSC stem from, we conduct two ablations. First, we include the translated summaries as standalone documents instead of prepending them to their English source documents. This tests whether the gains stem simply from adding informative non-English tokens, or from pairing those tokens with semantically aligned English text. Second, we train an English-only model in which each English document is prepended with its English summary, and evaluate it on English benchmarks. This tests whether summary conditioning itself, without translation, improves performance in a monolingual setting.



\subsection{To prepend or not to prepend}
\label{sec:abla-no-prepend}
\begin{figure*}[t]
    \centering
    \includegraphics[width=0.8\textwidth]{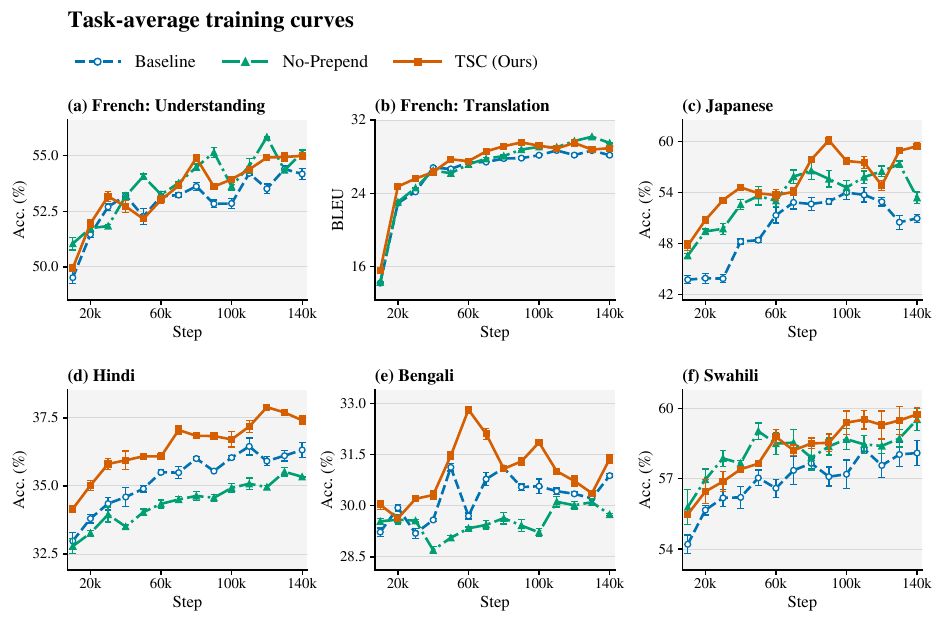}
    \caption{Task-average training curves of the baseline, No-Prepend, and TSC models through 140{,}000 steps.}
    \label{fig:abla_eval_average}
\end{figure*}

We train a \textit{No-Prepend} variant on the same data as TSC, in which the translated summaries are included as standalone documents rather than prepended to their English source documents. Figure~\ref{fig:abla_eval_average} compares the task-average learning curves of the baseline, the No-Prepend variant, and TSC; per-task curves are provided in Appendix~\ref{app:abla-curves}.

The effect of prepending depends strongly on the language. For Hindi and Bengali, TSC outperforms No-Prepend at nearly every checkpoint: adding translated summaries as short standalone documents does not help these languages and appears to hurt them (No-Prepend is below the baseline), whereas pairing the same summaries with their English sources yields consistent gains. For Japanese, No-Prepend improves over the baseline, but TSC achieves larger gains on JCommonsenseQA and XWinograd, while No-Prepend is stronger on JNLI. 
For French, TSC does not outperform No-Prepend on understanding tasks, possibly because French and English share both script and substantial vocabulary, so cross-lingual transfer occurs readily~\citep{pires-etal-2019-multilingual} regardless of how the French data is combined with English. For translation, however, TSC maintains a consistent advantage for French$\rightarrow$English throughout training, but not for English$\rightarrow$French, consistent with the French-to-English ordering of TSC examples discussed in \S\ref{sec:results}.
For Swahili, the two variants perform comparably and both clearly outperform the baseline; as Swahili is the lowest-resource language in our study, any additional Swahili text may be beneficial regardless of how it is incorporated into training. 

Overall, these results indicate that the gains of TSC do not stem solely from additional non-English tokens. Pairing translated summaries with their English sources is crucial for Hindi and Bengali and helps for specific tasks for Japanese and French, whereas for Swahili, the added summaries help regardless of pairing.

\subsection{English-only summaries}
\label{sec:abla-en-summary}
We train a model for which each English document is prepended with its LLM-generated English summary, without any non-English data. We compare it with a baseline trained on the same English documents without summaries, keeping the architecture, tokeniser, hyperparameters, and training budget of 140{,}000 steps fixed. As shown in Table~\ref{tab:tsc-140k-ablation-en}~(Appendix~\ref{app:abla-en}), the two models perform comparably on English benchmarks: 
The average differs by only $+0.35 \pm 0.34$ points, which is within seed variability.
%
Thus, prepending English summaries has no meaningful effect on English performance in our setting. This contrasts with the acceleration reported for metadata conditioning~\citep{gao2025metadata}, possibly because summaries restate the content of the document rather than providing additional information about its source.
Since this ablation is monolingual, however, it does not determine whether summary conditioning alone contributes to the non-English gains of TSC; we leave the corresponding multilingual control, prepending English summaries while keeping the native non-English data, to future work.

%% file: mi_part/mi_concept_readout.tex
\section{Cross-lingual Concept Readout}
\label{sec:mi-readout}

TSC yields its largest average gain over the baseline on Japanese (Section~\ref{sec:results}). The No-Prepend ablation (Section~\ref{sec:abla-no-prepend}) further shows that the incorporation of the translated summaries without prepending also matters. However, for Hindi and Bengali, No-Prepend falls below the baseline. To examine what TSC and No-Prepend change inside the models, we use cross-lingual concept patching~\citep{dumas-etal-2025-separating,korner-etal-2026-meanings}, a mechanistic interpretability method that measures how strongly a model's internal representations of a concept are shared across languages.



\begin{figure*}[t]
  \centering
  \includegraphics[width=\textwidth]{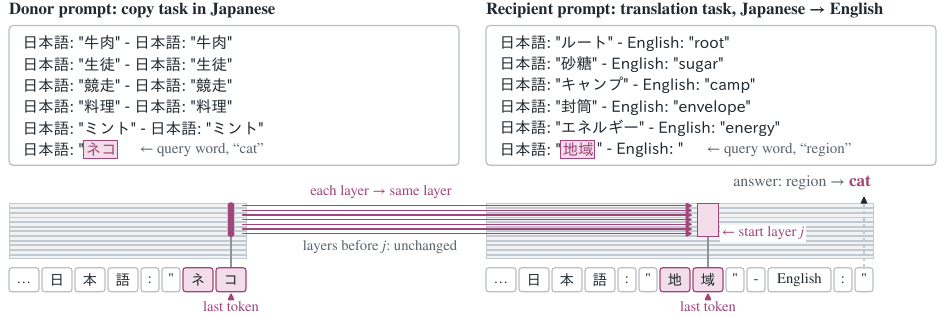}
  \caption{An example of concept patching on the bilingual English-Japanese model. \textbf{Top:} the donor prompt for a Japanese copy task, and the recipient prompt for a Japanese$\to$English translation task. Each contains five examples followed by a query word (highlighted): the Japanese word for \emph{cat} in the donor and the Japanese word for \emph{region} in the recipient. The donor prompt ends right after its query word, and the recipient prompt ends where the English answer begins. \textbf{Bottom:} the final tokens of each prompt (the examples are abbreviated as \ldots) and a schematic of the model's layers. From start layer $j$ onward, the hidden state at the last token of the recipient's word is replaced, layer by layer, with the donor's hidden state at the last token of its word; layers before $j$ are unchanged. When $j$ is early enough, the model's answer changes from \emph{region} to \emph{cat}.$^{*}$}
  \label{fig:mi-example}
  \par\vspace{3pt}
  \begin{minipage}{\textwidth}\footnotesize{$^{*}$The tokenizer has no token for the second character of the recipient's word, so it splits this character into three tokens, one for each byte of its UTF-8 encoding. The figure draws these three tokens as one box, and patching uses the last of them.}\end{minipage}
\end{figure*}

\paragraph{Concept patching}
This variant of activation patching tests whether the hidden state of a word encodes its meaning in a form the model can express in another language. Following~\citet{dumas-etal-2025-separating} and~\citet{korner-etal-2026-meanings}, we run the same model on two prompts: a \emph{donor} prompt, which asks the model to copy a word in the source language, and a \emph{recipient} prompt, which asks it to translate a different word from the source language into a target language. We then replace the hidden state at the last token of the recipient's word with the hidden state at the last token of the donor's word. If the model outputs the translation of the donor's word, it has recovered that word's meaning from the transferred state. As shown in an example in Figure~\ref{fig:mi-example}, the donor's query word is the Japanese word for \emph{cat} and the recipient's query word is the Japanese word for \emph{region}.


In implementation, we first sample donor--recipient pairs of English nouns from Multi-SimLex~\citep{vulic-etal-2020-multi}, a multilingual word-similarity dataset, using one English form per noun (e.g., \emph{cat} and \emph{region} in Figure~\ref{fig:mi-example}). We then obtain their Japanese, Hindi, and Bengali translations from the MUSE bilingual dictionaries~\citep{lample2018word}. These dictionaries often list several forms per noun, such as both the singular and plural of a Hindi noun, and we accept any of them. Appendix~\ref{app:mi-readout} gives the full procedure.

\begin{figure}[t]
  \centering
  \includegraphics[width=\columnwidth]{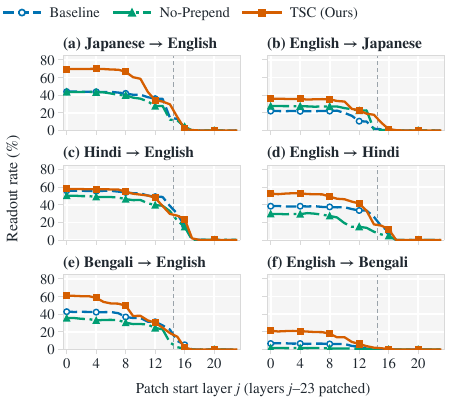}
  \caption{Readout rates of TSC, No-Prepend, and Baseline models for each start layer. The vertical grey line marks the end of the start layers averaged in Table~\ref{tab:mi-readout} (0--14).}
  \label{fig:mi-layers}
\end{figure}

\paragraph{Measures}
We analyse the models after 140{,}000 training steps, which have 24 Transformer layers, numbered from 0 (closest to the input) to 23. Cumulative patching with start layer $j$ (Figure~\ref{fig:mi-example}, bottom) replaces the hidden state at the last token of the recipient's word in layers $j$ through 23, each with the donor's hidden state from the same layer, and leaves earlier layers unchanged. For each language, we sample 256 prompt pairs with each of three random seeds, yielding 768 pairs. For each start layer, the \emph{readout rate} is the percentage of pairs whose output exactly matches one of the word forms of the donor concept in the output language, as shown in Figure~\ref{fig:mi-layers}. Readout succeeds only when patching starts early. 
%
To compare the model performances, we therefore summarise each model by its mean readout rate over start layers 0--14 as shown in Table~\ref{tab:mi-readout}, a range that ends before readout falls to near zero for most models and directions. \citet{korner-etal-2026-meanings} also found that concepts can be patched up to about layer 14 in a 24-layer model. 
Appendix~\ref{app:mi-readout} reports results for other end layers, from 7 to 16. All differences between models keep their sign, except the Hindi$\to$English difference between TSC and the baseline, which ranges from \ensuremath{-}0.5 to +1.9 points depending on the end layer.



\paragraph{Japanese}
For Japanese models, as shown in Table~\ref{tab:mi-readout}, Figure~\ref{fig:mi-layers}(a), and Figure~\ref{fig:mi-layers}(b), TSC outputs the donor concept in 58.8\% of Japanese$\to$English cases, against 39.8\% for the baseline and 37.2\% for No-Prepend when averaged over start layers 0--14. TSC's lead over No-Prepend is 21.6 points in Japanese$\to$English and 6.7 points in English$\to$Japanese. No-Prepend itself is 2.6 points slightly below the baseline in Japanese$\to$English and 6.1 points above it in English$\to$Japanese.

\input{mi_part/mi_table_main}

\paragraph{Hindi and Bengali}
For Hindi and Bengali, as discussed in Section~\ref{sec:abla-no-prepend}, our experiments have found that No-Prepend falls below the baseline on the downstream evaluations. As shown in Table~\ref{tab:mi-readout} and Figures~\ref{fig:mi-layers}(c)--(f), their readout rates are also lower than those of the baselines in all four Hindi and Bengali directions, by 4.1 to 12.2 points (Baseline$-$No-Prepend). TSC leads No-Prepend in all four directions, by 6.7 to 22.3 points (TSC$-$No-Prepend). TSC also outperforms the baseline in English$\to$Hindi and in both Bengali directions, by 8.8 to 10.1 points. In Hindi$\to$English, it is 0.4 points slightly below the baseline.

\paragraph{Interpretation}
A higher readout rate means that the model more often produces the donor concept in the other language after the hidden state of the donor's word is transferred. We use this as a functional measure of cross-lingual concept alignment. By this measure, TSC aligns word-level concepts between English and target non-English language more strongly. Its readout rate is higher than that of No-Prepend in all six directions and higher than that of the baseline in five. TSC and No-Prepend are trained on the same data, namely the original English and native non-English documents together with the translated summaries, but only TSC prepends each summary to its English source document. Since the two models differ only in this pairing, the stronger alignment observed in our concept patching analysis may stem from TSC seeing the same content in both languages within a single document during training.

%% file: mi_part/mi_table_main.tex

\begin{table}[t]
  \centering
  \small
  \setlength{\tabcolsep}{1.5pt}
  \renewcommand{\arraystretch}{1.0}
  \begin{tabular*}{\columnwidth}{@{\extracolsep{\fill}}lrrr@{}}
    \toprule
    Direction & Baseline & No-Prepend & TSC \\
    \midrule
    Japanese$\to$English & 39.8 & 37.2 & \textbf{58.8} \\
    English$\to$Japanese & 18.6 & 24.7 & \textbf{31.4} \\
    \addlinespace[3pt]
    Hindi$\to$English & \textbf{52.7} & 45.6 & 52.3 \\
    English$\to$Hindi & 36.3 & 24.1 & \textbf{46.4} \\
    \addlinespace[3pt]
    Bengali$\to$English & 37.3 & 29.7 & \textbf{46.1} \\
    English$\to$Bengali & 5.4 & 1.3 & \textbf{15.4} \\
    \bottomrule
  \end{tabular*}
  \caption{Readout rates of TSC, No-Prepend, and Baseline models. The highest rate in each row is in bold. 
  Appendix~\ref{app:mi-readout} discusses the low English$\to$Bengali rates.}
  \label{tab:mi-readout}
\end{table}

%% file: Experiments/tables/model-parameters.tex
\begin{table}[!t]
\small 
\centering
\setlength{\tabcolsep}{8pt}
\begin{tabular}{lc}
\toprule
 Hyperparameter &  Value\\
\midrule
Sequence Length &   2048 \\
    Number of Layers &    24 \\
   Embedding Size &    2048 \\
   FFN Hidden Size &   5504 \\
   Number of Heads &   16 \\
   Position Encodings &   RoPE \\
   Activation Function &   SwiGLU\\
   Layer Norm &   RMSNorm \\
   Learning Rate &   2E-4 \\
   Batch Size &   2048 \\
   Vocabulary Size &    32000 \\
 \midrule
   Embedding Parameters &   0.13B \\
   Non-Embedding Parameters &   1.21B\\
   Total Parameters &   1.34B\\
\bottomrule
\end{tabular}
\caption{Model and pretraining hyperparameters.} 
\label{tab:model-arch}
\end{table}

%% file: Experiments/tables/benchmarks.tex
\begin{table*}[t]
\centering
\footnotesize
\setlength{\tabcolsep}{3pt}
\begin{tabular}
{@{}p{0.30\textwidth}|p{0.43\textwidth}|p{\dimexpr0.27\textwidth-12pt-2\arrayrulewidth\relax}@{}}
\toprule
\textbf{Benchmark} & \textbf{Description} & \textbf{Languages evaluated} \\
\midrule
\multicolumn{3}{@{}l@{}}{\textbf{\textit{Commonsense reasoning}}} \\
CommonsenseQA (CSQA)~\cite{talmor-etal-2019-commonsenseqa,kurihara-etal-2022-jglue,nahin2025titullms} & Everyday commonsense knowledge. & Bengali, Japanese \\
XCOPA~\cite{ponti2020xcopa} & Causal commonsense reasoning (extension of COPA~\cite{roemmele2011choice}). & Swahili \\
PIQA~\cite{Bisk2020} & Physical commonsense reasoning. & English \\
XStoryCloze~\cite{lin2021few} & Plausible story endings. & English, Hindi, Swahili \\
HellaSwag~\cite{zellers2019hellaswag,lai2023okapi} & Commonsense sentence completion. & Bengali, English, French, Hindi \\
XWinograd~\cite{tikhonov2021s,muennighoff2022crosslingual} & Commonsense pronoun resolution (extension of the Winograd Schema Challenge~\cite{levesque2012winograd}). & English, Japanese \\
\midrule
\multicolumn{3}{@{}l@{}}{\textbf{\textit{Sentence-pair understanding}}} \\
XNLI~\cite{conneau2018xnli} & Multilingual entailment classification. & English, French \\
JNLI~\cite{kurihara-etal-2022-jglue} & Japanese entailment classification. & Japanese \\
PAWS~\cite{paws2019naacl} / PAWS-X~\cite{pawsx2019emnlp} & Paraphrase identification. & English, French \\
\midrule
\multicolumn{3}{@{}l@{}}{\textbf{\textit{Reading comprehension}}} \\
BoolQ~\cite{clark2019boolq,nahin2025titullms} & Yes/no questions over a passage. & Bengali \\
\midrule
\multicolumn{3}{@{}l@{}}{\textbf{\textit{Knowledge-based question answering}}} \\
ARC~\cite{clark2018think,lai2023okapi} & Grade-school science questions. ARC-C refers to the challenge set, ARC-E refers to the easy set. & Bengali, English, French, Hindi \\
\midrule
\multicolumn{3}{@{}l@{}}{\textbf{\textit{Vocabulary and grammar}}} \\
Vocab$^\dagger$~\cite{faysse2024croissantllm} & French vocabulary questions. & French \\
Grammar$^\dagger$~\cite{faysse2024croissantllm} & French grammar questions. & French \\
\midrule
\multicolumn{3}{@{}l@{}}{\textbf{\textit{Translation}}} \\
WMT14 En--Fr~\cite{bojar2014findings} & English$\rightarrow$French translation. & English, French \\
WMT14 Fr--En~\cite{bojar2014findings} & French$\rightarrow$English translation. & English, French \\
\bottomrule
\end{tabular}
\caption{Benchmarks used in our evaluation and the languages evaluated for each. Non-English versions of CommonsenseQA are JCommonsenseQA (JCSQA)~\cite{kurihara-etal-2022-jglue} for Japanese and CommonsenseQA Bangla~\cite{nahin2025titullms} for Bengali. The Bengali version of BoolQ~\cite{clark2019boolq} is BoolQ Bangla~\cite{nahin2025titullms}, a new dataset in the same format rather than a translation, pairing passages from Bangla Wikipedia, Banglapedia, and news articles with yes/no questions and answers generated by GPT-4. $^\dagger$FrenchBench tasks.}
\label{tab:benchmark_categories}
\end{table*}

%% file: Experiments/tables/eval-results-EN.tex
\begin{table*}[t]
\centering
\small
\setlength{\tabcolsep}{1.5pt}
\renewcommand{\arraystretch}{1.0}
\begin{tabular*}{\textwidth}{@{\extracolsep{\fill}}llrrrrrrrr@{}}
\toprule
Group & Model & ARC-E & ARC-C & HellaSwag & PIQA & PAWS-X & XNLI & XWinograd & \textbf{Average} \\
\midrule
\textbf{French} & Baseline & \tscscore{71.32}{0.40} & \tscscore{36.26}{0.62} & \tscbest{42.84}{0.16} & \tscbest{71.73}{0.17} & \tscbest{57.52}{0.75} & \tscbest{48.14}{0.44} & \tscbest{78.38}{0.16} & \tscbest{58.03}{0.25} \\
 & TSC (Ours) & \tscbest{71.90}{0.18} & \tscbest{37.12}{0.31} & \tscscore{42.34}{0.13} & \tscscore{71.60}{0.27} & \tscscore{53.38}{0.35} & \tscscore{47.48}{0.59} & \tscscore{77.48}{0.19} & \tscscore{57.33}{0.15} \\
 & $\Delta$ & \tscscore{+0.58}{0.54} & \tscscore{+0.85}{0.81} & \tscscore{-0.50}{0.26} & \tscscore{-0.13}{0.35} & \tscscore{-4.13}{1.04} & \tscscore{-0.66}{0.79} & \tscscore{-0.90}{0.26} & \tscscore{-0.70}{0.35} \\
\midrule
\textbf{Japanese} & Baseline & \tscscore{70.22}{0.25} & \tscscore{34.95}{0.75} & \tscscore{41.14}{0.12} & \tscscore{70.82}{0.13} & \tscbest{54.90}{0.46} & \tscscore{47.15}{0.25} & \tscscore{75.37}{0.49} & \tscscore{56.36}{0.12} \\
 & TSC (Ours) & \tscbest{70.26}{0.13} & \tscbest{35.01}{0.32} & \tscbest{41.83}{0.12} & \tscbest{71.20}{0.35} & \tscscore{53.12}{0.89} & \tscbest{47.83}{0.80} & \tscbest{77.61}{0.15} & \tscbest{56.69}{0.20} \\
 & $\Delta$ & \tscscore{+0.04}{0.34} & \tscscore{+0.06}{0.98} & \tscscore{+0.68}{0.03} & \tscscore{+0.38}{0.25} & \tscscore{-1.78}{0.45} & \tscscore{+0.68}{0.91} & \tscscore{+2.24}{0.34} & \tscscore{+0.33}{0.19} \\
\midrule
\textbf{Hindi} & Baseline & \tscscore{72.24}{0.60} & \tscbest{39.53}{0.52} & \tscbest{43.09}{0.21} & \tscscore{73.10}{0.25} & \tscscore{56.18}{1.82} & \tscscore{45.80}{1.21} & \tscscore{78.42}{0.91} & \tscscore{58.34}{0.16} \\
 & TSC (Ours) & \tscbest{72.33}{0.84} & \tscscore{39.16}{0.60} & \tscscore{43.05}{0.14} & \tscbest{73.18}{0.36} & \tscbest{57.63}{1.04} & \tscbest{47.30}{0.83} & \tscbest{78.59}{0.27} & \tscbest{58.75}{0.27} \\
 & $\Delta$ & \tscscore{+0.10}{1.35} & \tscscore{-0.37}{0.91} & \tscscore{-0.04}{0.27} & \tscscore{+0.07}{0.55} & \tscscore{+1.45}{0.88} & \tscscore{+1.50}{0.85} & \tscscore{+0.17}{0.72} & \tscscore{+0.41}{0.14} \\
\midrule
\textbf{Bengali} & Baseline & \tscbest{72.92}{0.09} & \tscbest{38.14}{0.17} & \tscbest{43.23}{0.12} & \tscscore{72.29}{0.06} & \tscbest{58.28}{1.48} & \tscscore{45.72}{0.82} & \tscbest{78.70}{0.78} & \tscbest{58.47}{0.31} \\
 & TSC (Ours) & \tscscore{71.86}{0.28} & \tscscore{34.84}{1.00} & \tscscore{42.88}{0.03} & \tscbest{73.12}{0.20} & \tscscore{56.83}{0.55} & \tscbest{45.92}{0.28} & \tscscore{77.18}{0.17} & \tscscore{57.52}{0.21} \\
 & $\Delta$ & \tscscore{-1.07}{0.25} & \tscscore{-3.30}{1.07} & \tscscore{-0.34}{0.13} & \tscscore{+0.83}{0.26} & \tscscore{-1.45}{1.35} & \tscscore{+0.20}{1.04} & \tscscore{-1.52}{0.60} & \tscscore{-0.95}{0.39} \\
\midrule
\textbf{Swahili} & Baseline & \tscbest{73.41}{0.24} & \tscbest{39.62}{0.36} & \tscbest{44.17}{0.18} & \tscbest{73.29}{0.52} & \tscscore{55.20}{0.38} & \tscscore{46.51}{0.43} & \tscbest{80.36}{0.37} & \tscbest{58.94}{0.16} \\
 & TSC (Ours) & \tscscore{72.88}{0.05} & \tscscore{39.02}{0.27} & \tscscore{43.71}{0.11} & \tscscore{72.76}{0.11} & \tscbest{56.03}{0.30} & \tscbest{47.10}{1.06} & \tscscore{77.94}{0.32} & \tscscore{58.49}{0.10} \\
 & $\Delta$ & \tscscore{-0.53}{0.19} & \tscscore{-0.60}{0.48} & \tscscore{-0.46}{0.21} & \tscscore{-0.53}{0.63} & \tscscore{+0.83}{0.59} & \tscscore{+0.59}{0.71} & \tscscore{-2.42}{0.16} & \tscscore{-0.45}{0.22} \\
\bottomrule
\end{tabular*}
\caption{English evaluation of the five bilingual model groups at 140{,}000 steps.
Entries report accuracy (\%), mean $\pm$ sample SD over three evaluation seeds.
Average equally weights these seven tasks within each run; its SD is computed across the three run averages.
}
\label{tab:tsc-140k-english}
\end{table*}

%% file: Experiments/tables/ablation-en.tex
\begin{table}[t]
\centering
\small
\setlength{\tabcolsep}{1.5pt}
\renewcommand{\arraystretch}{1.0}
\begin{tabular*}{\columnwidth}{@{\extracolsep{\fill}}lrrr@{}}
\toprule
Task & Baseline & TSC (en) & $\Delta$ \\
\midrule
\multicolumn{4}{@{}l}{\textit{English evaluation (Acc.)}} \\
ARC-E & \tscscore{74.48}{0.12} & \tscbest{74.69}{0.32} & \tscscore{+0.21}{0.42} \\
ARC-C & \tscbest{39.08}{0.60} & \tscscore{38.20}{0.13} & \tscscore{-0.88}{0.57} \\
HellaSwag & \tscscore{44.18}{0.25} & \tscbest{44.27}{0.31} & \tscscore{+0.09}{0.10} \\
PIQA & \tscscore{72.87}{0.35} & \tscbest{73.01}{0.05} & \tscscore{+0.15}{0.31} \\
PAWS-X & \tscscore{55.80}{0.85} & \tscbest{56.58}{0.51} & \tscscore{+0.78}{0.85} \\
XNLI & \tscscore{47.84}{0.64} & \tscbest{50.09}{0.39} & \tscscore{+2.25}{0.72} \\
XStoryCloze & \tscbest{68.52}{0.38} & \tscscore{68.34}{0.36} & \tscscore{-0.18}{0.68} \\
XWinograd & \tscscore{79.91}{0.73} & \tscbest{80.26}{0.39} & \tscscore{+0.34}{0.63} \\
\midrule
\textbf{Average} & \tscscore{60.34}{0.33} & \tscbest{60.68}{0.08} & \tscscore{+0.35}{0.34} \\
\bottomrule
\end{tabular*}
\caption{English-only ablation at 140{,}000 steps. \mbox{Acc.} (\%) is mean $\pm$ sample SD over three evaluation seeds.
}
\label{tab:tsc-140k-ablation-en}
\end{table}

%% file: mi_part/mi_appendix.tex
\section{Concept-Patching Details}
\label{app:mi-readout}

\paragraph{Concepts}
We define a concept as an English noun together with its translations into Japanese, Hindi, and Bengali. Following~\citet{korner-etal-2026-meanings}, we take the English words of the 1,888 word pairs in Multi-SimLex~\citep{vulic-etal-2020-multi}. As in their procedure, we drop the 18 words that Multi-SimLex labels as different parts of speech in different pairs, such as \emph{watch}, which is a noun in one pair and a verb in another. Their procedure also requires each word to have a translation in every Multi-SimLex language, which removes one more word, \emph{zero}. We then keep the nouns, yielding 1,290 concepts. We take the Japanese, Hindi, and Bengali translations from the machine-generated MUSE dictionaries~\citep{lample2018word} and keep every translation written in the script of its language, including spelling variants and synonyms. Of the 1,290 nouns, 1,078 have Japanese forms, 1,141 have Hindi forms, and 1,078 have Bengali forms.

\paragraph{Prompt pairs}
For each of Japanese, Hindi, and Bengali, we pair two different concepts that both have translations in that language, one as the donor and one as the recipient. As in~\citet{korner-etal-2026-meanings}, we take the first token of every word form of each concept, both the English noun and its translations, and pair two concepts only if they share no first token. For each random seed, we shuffle all such pairs and go through them in order until we have kept 256 pairs. We skip a pair if (1) its donor is already the donor of a kept pair, or (2) its recipient is already the recipient of a kept pair. For example, suppose that the shuffled list begins with the following four pairs, each written as (donor, recipient):
\begin{enumerate}\setlength{\itemsep}{0pt}
  \item (\emph{cat}, \emph{region}): kept, because neither concept has been used yet;
  \item (\emph{cat}, \emph{dog}): skipped, because \emph{cat} is already a donor;
  \item (\emph{dog}, \emph{region}): skipped, because \emph{region} is already a recipient;
  \item (\emph{dog}, \emph{cat}): kept, because \emph{dog} is not yet a donor and \emph{cat} is not yet a recipient.
\end{enumerate}
The three seeds are drawn independently. Each seed starts with no concept used, so the same concept can be a donor again in another seed. We keep all 768 pairs and do not remove repeats. The same pairs, with the same donor and recipient, serve both translation directions, for example Japanese$\to$English and English$\to$Japanese.

\paragraph{Bengali pairs}
In UTF-8, most Bengali letters start with the same two bytes, and the tokenizer of the Hindi and Bengali models has a single token for these two bytes. Of the 1,078 nouns, 1,039 have a Bengali form that begins with this token, and two such nouns cannot be paired. Every Bengali pair therefore contains at least one of the other 39 nouns, and only 6.8\% of the possible Bengali pairs meet the first-token condition (Japanese: 97.0\%, Hindi: 90.0\%). With the skip rule, fewer than 256 Bengali pairs fit. We therefore relax it step by step: a concept may be the donor of up to two kept pairs and the recipient of up to two, then up to three, and so on, until 256 pairs are kept. The 768 Bengali pairs thus contain only 374 different donor concepts (Japanese: 612, Hindi: 601), and the Bengali results rest on a narrower set of words than the Japanese and Hindi results.

\paragraph{Prompts}
For a direction from one language into another language, the recipient prompt is a translation task. It has five example lines and then a query line, after which the model writes its answer.
The donor prompt is a copy task and ends right after its query word.
Figure~\ref{fig:mi-example} in Section~\ref{sec:mi-readout} shows both prompts for Japanese$\to$English. 
For each prompt pair, we draw ten different concepts other than the two query concepts at random and give five to each prompt, so the two prompts share no example concept.

\paragraph{Patching}
The model has 24 Transformer layers, numbered 0 to 23. Each layer outputs one vector for every token, which is the token's hidden state in that layer, and passes it on to the next layer. We first run the donor prompt through the model once. The donor prompt ends immediately after its query word. For every layer, we save the hidden state at the last token of the donor prompt, which is the last token of the donor's query word. This gives one saved vector per layer. We then run the recipient prompt once for each start layer $j$, from 0 to 23. In each layer from $j$ to 23, we replace the hidden state at the last token of the recipient's query word with that layer's saved donor hidden state. If the query word consists of several tokens, we replace the hidden state only at its last token. We leave layers before $j$ and all other tokens of the recipient prompt unchanged. We make these replacements only while the model processes the recipient prompt. After that, the model generates its answer one token at a time, as described below. We do not replace any hidden state during this generation. 

\paragraph{Decoding and scoring}
Each recipient prompt ends with the opening quotation mark of the answer, as in \texttt{X: "query" - Y: "}, where X and Y name the input and output languages. After this prompt, we generate at most ten new tokens with greedy decoding, taking the most probable next token at each step. Normally, the model writes the answer, followed by a closing quotation mark \texttt{"}, a line break, and the beginning of the next example line. We take the generated text before the first \texttt{"} as the answer. If the generated text contains no \texttt{"}, which happens when the ten tokens run out before the model closes the answer, we take the whole generated text. We remove spaces and line breaks at the beginning and end of the answer, and we keep spaces inside the answer. We then lowercase both the answer and the word forms of the donor concept in the output language. We count the answer as a readout if it is then exactly equal to one of these word forms. For example, consider the TSC model on Japanese$\to$English with the donor concept \emph{cat} and the recipient concept \emph{region}. With start layer 0, the generated text is \texttt{cat"} followed by a line break and the beginning of the next example line. The answer is \texttt{cat}, which matches the donor concept. With start layer 16, the generated text begins with \texttt{region"}, so the answer is \texttt{region}, which does not match. In another example, the generated text \texttt{Oasis"} gives the answer \texttt{Oasis}, which matches the donor's word form \texttt{oasis} after lowercasing.

\begin{table*}[t]
  \centering
  \small
  \setlength{\tabcolsep}{1.5pt}
  \renewcommand{\arraystretch}{1.0}
  \begin{tabular*}{\textwidth}{@{\extracolsep{\fill}}l*{10}{r}@{}}
    \toprule
    & \multicolumn{10}{c}{End layer $k$} \\
    \cmidrule(l){2-11}
    Model & 7 & 8 & 9 & 10 & 11 & 12 & 13 & 14 & 15 & 16 \\
    \midrule
    \multicolumn{11}{@{}l}{\textit{Japanese$\to$English}} \\
    Baseline & 43.8 & 43.6 & 43.3 & 43.0 & 42.6 & 42.1 & 41.6 & 39.8 & 38.2 & 36.1 \\
    No-Prepend & 42.9 & 42.6 & 42.1 & 41.5 & 40.8 & 39.8 & 38.9 & 37.2 & 35.5 & 33.7 \\
    TSC & \textbf{69.5} & \textbf{69.2} & \textbf{68.3} & \textbf{67.4} & \textbf{65.5} & \textbf{63.1} & \textbf{60.9} & \textbf{58.8} & \textbf{56.1} & \textbf{52.9} \\
    \addlinespace[3pt]
    \multicolumn{11}{@{}l}{\textit{English$\to$Japanese}} \\
    Baseline & 21.9 & 21.9 & 22.0 & 21.8 & 21.4 & 20.6 & 19.8 & 18.6 & 17.5 & 16.6 \\
    No-Prepend & 27.5 & 27.4 & 27.4 & 27.2 & 27.0 & 26.6 & 26.3 & 24.7 & 23.2 & 21.8 \\
    TSC & \textbf{35.7} & \textbf{35.6} & \textbf{35.4} & \textbf{35.2} & \textbf{34.3} & \textbf{33.4} & \textbf{32.4} & \textbf{31.4} & \textbf{30.0} & \textbf{28.3} \\
    \addlinespace[3pt]
    \multicolumn{11}{@{}l}{\textit{Hindi$\to$English}} \\
    Baseline & 55.6 & 55.4 & 55.2 & 54.9 & 54.6 & 54.1 & 53.8 & \textbf{52.7} & \textbf{51.3} & \textbf{49.2} \\
    No-Prepend & 49.3 & 49.0 & 48.6 & 48.3 & 47.7 & 47.1 & 46.5 & 45.6 & 44.1 & 42.4 \\
    TSC & \textbf{57.5} & \textbf{57.2} & \textbf{56.7} & \textbf{56.3} & \textbf{55.8} & \textbf{55.2} & \textbf{54.0} & 52.3 & 50.8 & \textbf{49.2} \\
    \addlinespace[3pt]
    \multicolumn{11}{@{}l}{\textit{English$\to$Hindi}} \\
    Baseline & 38.1 & 38.0 & 37.9 & 37.9 & 37.6 & 37.3 & 37.0 & 36.3 & 35.0 & 33.5 \\
    No-Prepend & 29.5 & 29.3 & 28.9 & 28.0 & 26.9 & 26.0 & 25.1 & 24.1 & 23.0 & 21.9 \\
    TSC & \textbf{52.3} & \textbf{52.0} & \textbf{51.5} & \textbf{51.0} & \textbf{50.4} & \textbf{49.7} & \textbf{48.5} & \textbf{46.4} & \textbf{44.6} & \textbf{42.7} \\
    \addlinespace[3pt]
    \multicolumn{11}{@{}l}{\textit{Bengali$\to$English}} \\
    Baseline & 42.4 & 41.8 & 41.1 & 40.6 & 39.9 & 39.2 & 38.4 & 37.3 & 35.6 & 33.8 \\
    No-Prepend & 34.2 & 33.8 & 33.3 & 32.9 & 32.5 & 31.9 & 31.2 & 29.7 & 28.0 & 26.4 \\
    TSC & \textbf{57.4} & \textbf{56.6} & \textbf{54.7} & \textbf{53.2} & \textbf{51.5} & \textbf{49.9} & \textbf{48.1} & \textbf{46.1} & \textbf{44.1} & \textbf{41.7} \\
    \addlinespace[3pt]
    \multicolumn{11}{@{}l}{\textit{English$\to$Bengali}} \\
    Baseline & 6.8 & 6.7 & 6.6 & 6.6 & 6.4 & 6.1 & 5.8 & 5.4 & 5.1 & 4.8 \\
    No-Prepend & 1.6 & 1.6 & 1.5 & 1.5 & 1.5 & 1.4 & 1.4 & 1.3 & 1.2 & 1.1 \\
    TSC & \textbf{20.5} & \textbf{20.2} & \textbf{19.6} & \textbf{19.1} & \textbf{18.2} & \textbf{17.3} & \textbf{16.3} & \textbf{15.4} & \textbf{14.5} & \textbf{13.7} \\
    \bottomrule
  \end{tabular*}
  \caption{Mean readout rate (\%) over start layers 0 to $k$ for each end layer $k$ from 7 to 16. The highest rate of each direction and end layer is in bold. The column for $k = 14$ repeats Table~\ref{tab:mi-readout}.}
  \label{tab:mi-end-layers}
\end{table*}

\paragraph{End of the averaging range}
In Section~\ref{sec:mi-readout}, we summarise each model by its mean readout rate over start layers 0 to 14 (Table~\ref{tab:mi-readout}). For an end layer $k$, we compute the mean readout rate as the average of the $k+1$ readout rates at start layers 0 to $k$. Table~\ref{tab:mi-end-layers} lists this mean for every model and direction and for every $k$ from 7 to 16. Its column for $k = 14$ repeats Table~\ref{tab:mi-readout}. As shown in Figure~\ref{fig:mi-layers}, every mean is lower at $k = 16$ than at $k = 7$, because the readout rate drops at later start layers. For example, the mean for TSC in Japanese$\to$English is 69.5\% at $k = 7$, 58.8\% at $k = 14$, and 52.9\% at $k = 16$. In five of the six directions, the three models keep the same order for every $k$ from 7 to 16. The exception is Hindi$\to$English: TSC is above the baseline for $k$ from 7 to 13, slightly below it for $k = 14$ and $k = 15$, and equal to it at $k = 16$ (49.2\% each). No-Prepend is below both for every $k$. 
Overall, TSC is above No-Prepend in all six directions for every $k$ from 7 to 16. TSC is above the baseline in all six directions for $k$ from 7 to 13. For $k$ from 14 to 16, it is above the baseline in five directions, with Hindi$\to$English as the exception.

\paragraph{English$\to$Bengali}
For English$\to$Bengali, the readout rates of all three models are low, and many answers are cut off by the ten-token limit. Over the start layers 0--14, the generated text has no closing quotation mark in 39.8\%, 30.3\%, and 32.2\% of the patched answers for the baseline, No-Prepend, and TSC models, against at most 6.0\% in the other directions.